\documentclass[sigplan,screen]{acmart}
\usepackage[symbol]{footmisc}
\AtBeginDocument{%
  \providecommand\BibTeX{{%
    \normalfont B\kern-0.5em{\scshape i\kern-0.25em b}\kern-0.8em\TeX}}}

\copyrightyear{2022}
\acmYear{2022}
\setcopyright{acmcopyright}\acmConference[SETN 2022]{12th Hellenic Conference on Artificial Intelligence}{September 7--9, 2022}{Corfu, Greece}
\acmBooktitle{12th Hellenic Conference on Artificial Intelligence (SETN 2022), September 7--9, 2022, Corfu, Greece}
\acmPrice{15.00}
\acmDOI{10.1145/3549737.3549780}
\acmISBN{978-1-4503-9597-7/22/09}

\begin{document}

\title{Multi-layer Representation Learning for Robust OOD Image Classification}

\author{Aristotelis Ballas}
\authornote{Both authors contributed equally to this research.}
\orcid{0000-0003-1683-8433}
\affiliation{%
  \institution{Department of Informatics and Telematics}
  \institution{Harokopio University}
  \streetaddress{Omirou 9, Tavros}
  \city{Athens}
  \country{Greece}
}
\email{aballas@hua.gr}

\author{Christos Diou}
\orcid{0000-0002-2461-1928}
\authornotemark[1]
\affiliation{%
  \institution{Department of Informatics and Telematics}
  \institution{Harokopio University}
  \streetaddress{Omirou 9, Tavros}
  \city{Athens}
  \country{Greece}}
\email{cdiou@hua.gr}

\renewcommand{\shortauthors}{Ballas and Diou.}

\begin{abstract}
Convolutional Neural Networks have become the norm in image classification.
Nevertheless, their difficulty to maintain high accuracy across datasets has become apparent
in the past few years. In order to utilize such models in real-world scenarios and
applications, they must be able to provide trustworthy predictions on unseen data. In this
paper, we argue that extracting features from a CNN's intermediate layers can assist in the
model's final prediction. Specifically, we adapt the Hypercolumns method to a ResNet-18
and find a significant increase in the model's accuracy, when evaluating on the NICO dataset.
\end{abstract}



\maketitle

\section{Introduction}
In the past few years, Deep Learning has established itself in academia and industry.
In particular, Convolutional Neural Networks have dominated image
classification \cite{krizhevsky_imagenet_2012}, achieving near-human, if not superhuman
\cite{DBLP:journals/corr/HeZR015}, accuracy.
Despite their outstanding results in IID (independent and identically distributed) datasets, most models today fail to generalize well on
unseen or out of distribution (OOD) settings \cite{pmlr-v97-recht19a}, since they tend to incorporate statistical correlations
present in the training data \cite{arjovsky_invariant_2020}. In real-world scenarios, data hardly ever originate from the
same distribution, creating a need for approaches that are able to distinguish between biases and trivial features and make decisions
based on invariant factors. This has been a long standing issue in the
Machine Learning community and, as a result, in 2011 the {\itshape Domain Generalization} (DG) problem \cite{blanchard_generalizing_2011} was formally introduced.
In the DG setting, the data used to evaluate the trained model originate from
a different distribution than the training data. By making predictions on unseen data distributions, we
can appropriately assess the model's ability to generalize.


\begin{figure}[t]
  \centering
  \includegraphics[width=\linewidth]{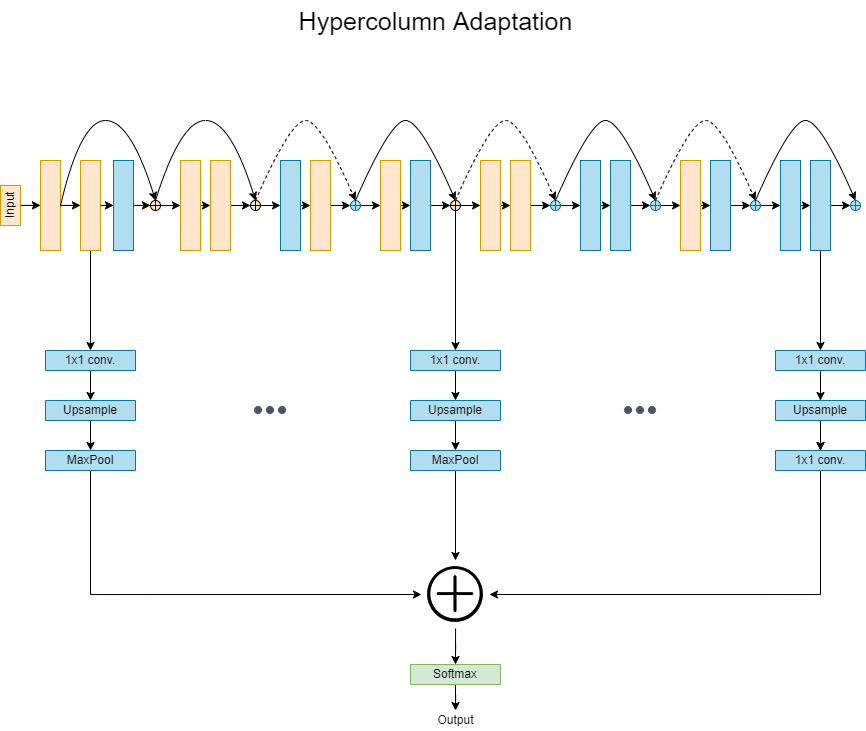}
  \caption{Visualization of the proposed model. The backbone of the architecture
  is a vanilla ResNet-18. By extracting feature maps from intermediate outputs of convolutional layers in the backbone model,
  our method takes advantage of the CNN's ability to detect edge and bar-like figures in the early
  layers and combines them with the semantic features extracted from layers further down the network. We argue that
  by incorporating outputs from different intermediate levels of the network, we enable the model
  to disentangle the invariant qualities of an object. We extract features from the layers and residual connections marked in
  light blue.}
  \Description{Visualization of the proposed model.}
  \label{model}
\end{figure}

In this work, we propose a method to tackle DG by adapting Hypercolumns
\cite{hariharan_hypercolumns_2015} to extract local attributes of an image in earlier layers and semantics in layers further down
the architecture. We hypothesize that by incorporating features from across the network, a classifier can be trained to ignore spurious correlations in the dataset and make predictions based on
invariant features. We evaluate our method by experimenting on NICO \cite{he2021towards}, a dataset specifically designed for OOD image
classification and are able to achieve promising results. Furthermore, to confirm our assumptions and intuition, we provide visual evidence
of our model's ability to distinguish between spurious and invariant characteristics present in an image.

\subsection{Domain Generalization}
In this section, we formally introduce the notations and definitions of DG.
Let $X$ be an input (feature) space and $Y$ an output (label) space. A domain is defined
as a joint distribution $\mathcal{P(X,Y)}$ $\sim$ $\mathcal{P_{XY}}$ on $\mathcal{X \times Y}$ .

In DG, the training and test distributions are OOD, in the sense that we are given S source (training) domains and T target (test) domains,
where $\mathcal{P}^i_\mathcal{XY}$ $\neq$ $\mathcal{P}^j_\mathcal{XY}$, 1 $\leq$ i, j $\leq$ S, T. Given labeled source domains S, the goal is to learn a model $\mathcal{F}$ ,
trained on data from S, which can adequately generalize to an unseen domain T.

\section{Related Work}
Domain Generalization is arguably one of the most challenging problems in Machine Learning. To this end,
a plethora of approaches have been proposed in the past few years. The most closely related fields to DG
are:
\begin{itemize}
    \item {\itshape Domain Adaptation} \cite{wang_deep_2018} and
{\itshape Transfer Learning} \cite{zhuang_comprehensive_2021} methods ,
which are perhaps the most common, focus on boosting their accuracy on unseen data by fine-tuning pretrained models on the target domain(s).
    \item {\itshape Meta-Learning} \cite{huisman_survey_2021}, aims to learn-to-learn and select the best method for solving the issue at hand.
    \item {\itshape Continual Learning} \cite{mai_online_2022} algorithms are used to overcome the issue of catastrophic forgetting by remembering the knowledge acquired over time and domains.
    \item {\itshape Zero-Shot Learning} \cite{Xian_2017_CVPR}, like DG deals with unseen distributions but in the label space.
\end{itemize}

With regard to learning stable or invariant features across domains, besides the baseline CNN proposed in the original paper \cite{he2021towards}, several other methods have been suggested.
To address the issues of complex, non-linear correlations between data in DG, the authors
of \cite{zhang2021deep} propose StableNet, a model which utilizes Random Fourier Features for sample weighting.
In \cite{Li_2021_CVPR}, the authors introduce the LIRR algorithm in the Semi-Supervised Domain Adaptation setting, for learning invariant representations and risks.
Another approach is to use gradient-based semantic augmentation \cite{bai2020decaug} to improve the generalizability of a model. Finally, the causal structure of the data can also be
utilized while a model is trained, as shown in \cite{sun2021latent}.

\section{Methodology}

\subsection{Hypercolumns}
Hypercolumns were first introduced in Neuroscience by Hubel and Wiesel \cite{hubel_receptive_1962},
in order to describe a vertical set of V1 neurons that behave similarly
to optical stimuli of the retina. The authors of \cite{hariharan_hypercolumns_2015}
borrowed this term and applied its fundamental attributes to a
Convolutional Neural Network, in an attempt to leverage the different
levels of information passed on the network's intermediate layers. Namely,
a Hypercolumn at a certain location is a stacked vector of the layer
outputs of the CNN's units above said location. In order to classify pixels using
Hypercolumns, it is assumed that bounding boxes of the points of interest have
been provided from an object detection system. For each bounding box,
a ${50 \times 50}$ heatmap (locations) is predicted, which is projected onto
the initial image and then passed into a CNN. Selected intermediate outputs of
the CNN are then concatenated into a vector and each location is classified
via ${1 \times 1}$ convolutional and fully connected layers. In the
same paper, the Efficient Hypercolumn method is also described, where the
${1 \times 1}$ convolutions are replaced by ${n \times n}$
convolutions and upsampling. In a subsequent work \cite{7486965}, the
authors were able to significantly  speed up their Hypercolumn pipeline by passing the whole image through the
CNN and cropping the bounding box locations afterwards.

Hypercolumns have been used for semantic segmentation
\cite{7486965}, object detection
\cite{bochkovskiy_yolov4_2020},
visual correspondence \cite{Min_2019_ICCV} and in some
cases for abnormality detection in the Biomedical domain
\cite{togacar_enhancing_2021}.
However, all above implementations construct their hypercolumns by concatenating
the upsampled images into a hyper vector, possibly without taking full advantage of
the already extracted features of the original image.
Thus, we propose a novel implementation of the original Hypercolumn method and adapt it
for robust image classification in the DG setting.

\subsection{Adapting hypercolumns to robust image classification}
Due to their convolutional nature, earlier layers in a CNN are prominent in detecting edges and bars, but cannot
distinguish between edges that belong to a vehicle or an animal, per se. The extracted information is then passed down the network and generalized
in the final layer, where inference occurs. An object or class consists of {\itshape features} which remain {\itshape invariant}
\cite{arjovsky_invariant_2020} across domains. For example, a horse still has legs and a mane, whether it is standing aside a person, lying in sand
or galloping through snow. Therefore, by disentangling an input image into distinct features, a causal decision can be made based upon the features present in the image.
Following the above example, the presence of `legs' in an image lead us to believe that an animal is most likely depicted. The presence of a
`mane', `long tail' and `oval-shaped' hoove make us confident that the animal is a horse. We argue that by taking advantage of the early and
intermediate features of a CNN, we can `push' a model to learn these invariant features (i.e edges and bars which correspond to \textit{parts} of the depicted objects).
For our model, we follow the original Hypercolumns implementation and select the outputs from intermediate conv layers, pass them through a ${1 \times 1}$ conv
layer and then upsample the outputs via bilinear interpolation. However, before concatenating the upsampled images to a hyper vector, we pass
them through a ${56 \times 56}$ MaxPool2D layer, with ${26 \times 26}$ strides, in an attempt to capture the features of the depicted class. After
the pooling layer, the outputs are concatenated into a hyper vector and passed through a classification head, which consists of a fully connected
layer followed by a softmax activation. Our model architecture is depicted in Figure 1.

\section{Experiments}

\subsection{Datasets}
In our experiments we adopt the \textbf{NICO} \cite{he2021towards} dataset. The NICO dataset was created for OOD image classification and is therefore a good
starting point for the evaluation of the robustness of classification algorithms. NICO contains 2 super-classes of \textit{Animal} and \textit{Vehicle}. The \textit{Animal}
superclass contains 10 classes and the \textit{Vehicle} contains 9. Each class contains 9 or 10 \textit{contexts}, which try to simulate real world scenarios, such as
`airplane aside mountain' and `airplane on grass'. The total images in the dataset are 25.000.

\subsection{Experimental Setup}
To evaluate our model we follow the \emph{leave-one-domain-out} protocol as described in \cite{Li_2017_ICCV}, where in our
case a domain is a context of a class. In order to demonstrate the robustness of our model, during training we select to hold out 3, 5 and 7
contexts from each class. For the backbone of our model, we select a ResNet-18, pre-trained on ImageNet. The selected
intermediate layers include all residual half block layers (i.e., those leading to reduction of the output size), as well as selected
convolutional layers. All later layers of the network are included. We train the model with
SGD for 30 epochs and with a batch size of 32 images. The learning rate is initially set at 0.001 and decays with a rate of of 0.1 at epoch 24. For
a baseline, we used a vanilla pre-trained ResNet-18 with the same hyperparameters as above. Both models were implemented with PyTorch on one NVIDIA RTX A5000 GPU.

\begin{table}
  \caption{Top-1\% Accuracy Results on the \textbf{NICO} Dataset when leaving out N contexts. If an image class has C contexts, we create a
  training split with the $\mathbf{C-N}$ contexts and evaluate our model on the remaining N. The presented results are averaged over 3 runs.}
  \label{tab:commands}
  \begin{tabular}{ccccl}
    \toprule
        Model & N=3 & N=5 & N=7 \\
    \midrule
    \texttt ResNet-18 & 79.6 & 78.1 & 78.9 \\
    \texttt Our Model & \textbf{82.8} & \textbf{81.3} &  \textbf{83.0}\\
    \bottomrule
  \end{tabular}
\end{table}

\begin{figure}
\centering
\includegraphics[width=.9\linewidth]{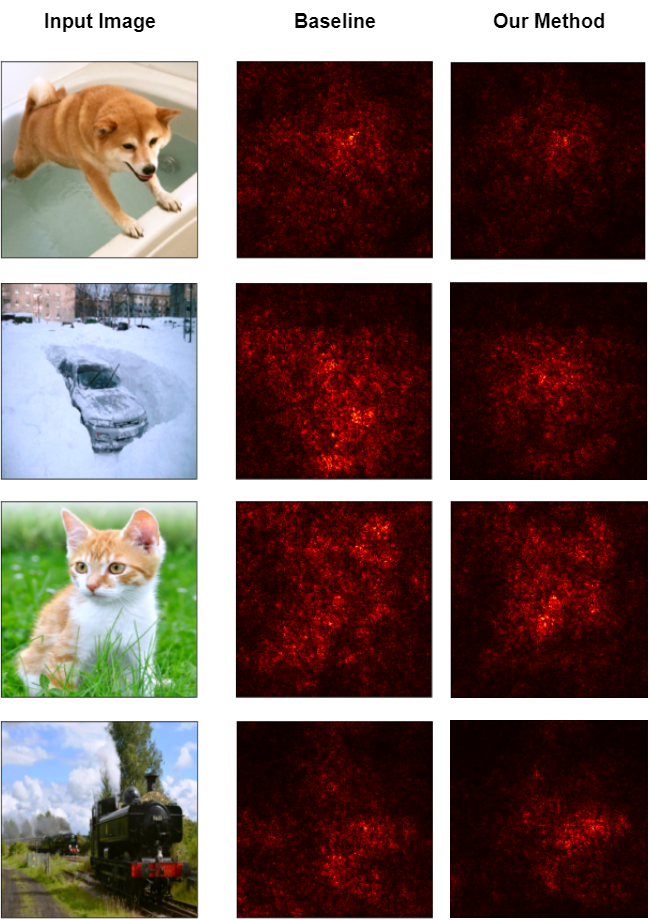}
\caption{Visualization of saliency maps, produced by the baseline vanilla ResNet-18 model and our method, from images in the NICO dataset.
The brighter the pixel, the more it contributes to the model's prediction. To a certain degree, our approach disregards the pixels corresponding
to the contexts and spurious correlations (i.e water, snow, grass and road) in each image, and focuses on the depicted object.}
\label{fig:saliency}
\end{figure}

\subsection{Results}
Table 1 depicts the averaged results on NICO after 3 runs. We can observe that our model outperforms the baseline by approximately 3.2\% when 3
contexts are left out, 3.2\% in the case where 5 contexts are left out and 2.4\% when we leave out 7.

To validate our initial assumptions, we also visualize our model's prediction with saliency maps. To be more specific, we adopted the
\emph{Image-Specific Class Saliency} method, as proposed in \cite{simonyan2014deep}. By computing and visualizing the gradient of the loss
function for the predicted class, with respect to the input pixels, one is able to produce a map of the pixels affecting the model's prediction.
As shown in Fig. \ref{fig:saliency}, the brightness in the saliency maps indicate the pixels which the model pays more `attention' to. Due to
spurious correlations present in the training data, the vanilla ResNet-18 tends to make assumptions based on unimportant features of the image
(e.g. water, snow, grass and road - indicated by the bright pixels around the object in the saliency maps), while our method seems to infer based
on features of the object itself and to some extent overlook the context features.

\section{Conclusion}
In this paper we attempt to tackle the Domain Generalization problem by adopting the Hypercolumns method and adapting it for robust image
classification. We argue that by utilizing the extracted features from a CNN's intermediate layers, the model can be forced to focus on the
invariant features in an image. This claim is supported by the results of our experiments on NICO, a dataset dominated by spurious correlations,
where we demonstrate our model's ability to perform well on unseen data. Through visual examples, we show that our method is capable of emphasizing
on the causal characteristics of an object and not on the inconsequential features of the input image. As future work, we aim to advance our method's
extraction mechanism and conduct further experiments on additional datasets.

\begin{acks}
The work leading to these results has received funding from the European Union’s Horizon 2020 research and innovation programme under Grant Agreement No. 965231
project REBECCA (REsearch on BrEast Cancer induced chronic conditions supported by Causal Analysis of multi-source data).
\end{acks}

\bibliographystyle{ACM-Reference-Format}
\bibliography{library}

\end{document}